\documentclass[runningheads]{llncs}
\usepackage{graphicx}
\usepackage{amsmath,amssymb} % define this before the line numbering.
\usepackage{color}
\usepackage{stfloats}
\usepackage{tablefootnote}
\usepackage{amsmath}
\usepackage{amssymb}
\usepackage{booktabs}
\usepackage{array}
\usepackage{caption}
\usepackage[pagebackref=true,breaklinks=true,letterpaper=true,colorlinks,bookmarks=false]{hyperref}

\begin{document}
% \renewcommand\thelinenumber{\color[rgb]{0.2,0.5,0.8}\normalfont\sffamily\scriptsize\arabic{linenumber}\color[rgb]{0,0,0}}
% \renewcommand\makeLineNumber {\hss\thelinenumber\ \hspace{6mm} \rlap{\hskip\textwidth\ \hspace{6.5mm}\thelinenumber}}
% \linenumbers
\pagestyle{headings}
\mainmatter

\title{Penalizing Top Performers: Conservative Loss for Semantic Segmentation Adaptation} % Replace with your title
%\title{Penalizing Top Performers: Gradient Reversal Loss for Urban Scene Adaptation}

\titlerunning{Penalizing Top Performers: Conservative Loss}

\authorrunning{X. Zhu, H. Zhou, C Yang, J. Shi, D. Lin}

\author{Xinge Zhu$^{\dag}$, Hui Zhou$^{\S}$, Ceyuan Yang$^{\dag}$, Jianping Shi$^{\S}$, Dahua Lin$^{\dag}$}
\institute{$^{\dag}$CUHK-SenseTime Joint Lab, CUHK\\$^{\S}$SenseTime Research \\ \email{\{zx018,dhlin\}@ie.cuhk.edu.hk\\ \{zhouhui,yangceyuan,shijianping\}@sensetime.com} }

\maketitle

\begin{abstract}
Due to the expensive and time-consuming annotations (e.g., segmentation) for real-world images, recent works in computer vision resort to synthetic data. However, the performance on the real image often drops significantly because of the domain shift between the synthetic data and the real images. In this setting, domain adaptation brings an appealing option. The effective approaches of domain adaptation shape the representations that (1) are discriminative for the main task and (2) have good generalization capability for domain shift. 
To this end, we propose a novel loss function, i.e., Conservative Loss, which penalizes the extreme good and bad cases while encouraging the moderate examples. More specifically, it
enables the network to learn features that are discriminative by gradient descent and are invariant to the change of domains via gradient ascend method. 
%In particular, the gradient ascend penalizes the case that the model performs well on training data (source data) but poorly on test data (target data). 
Extensive experiments on synthetic to real segmentation adaptation show our proposed method achieves state of the art results. Ablation studies give more insights into properties of the Conservative Loss.
Exploratory experiments and discussion demonstrate that our Conservative Loss has good flexibility rather than restricting an exact form.
%Additional exploratory experiments and discussion demonstrate that ($\text{{\romannumeral1}}$) our Conservative Loss  has good scalability and ($\text{{\romannumeral2}}$) our model is able to effectively generalize to Semi-Supervised Learning. 
% the model overfits to source domain data.
% domain generalization or domain invariant representations
% 1): learn to shape discriminative features
% 2): have good capability of generalization 
% the method is to escape the overfitting of the source domain
%\keywords{semantic segmentation, unsupervised domain adaptation, synthetic to real}
\end{abstract}

\section{Introduction}
\label{intro}
%large annotations, expensive, synthetic data from computer graphics
%
%
Deep convolutional neural networks have brought impressive advances to the state of the art across a multitude of tasks in computer vision~\cite{PSP,vgg16,FCN}. At the same time, these significant leaps require a large amount of labeled data. For some pixel-level tasks, e.g., semantic segmentation, obtaining a fine-grained label is expensive and time-consuming. In~\cite{cityscapes}, they report that it takes more than 90 minutes for manually labeling a single image. Recent advances in Computer Graphics~\cite{play} offer an alternative solution to address the data issue. In~\cite{play}, they automatically capture both images and fine-grained labels from GTAV game with the speed faster than human in several orders of magnitude.  

% domain shift, mismatch distribution, (tend to overfit to synthic images), domain changes. Some domain approaches.
However, models trained on the synthetic data fail to perform well on the real-world images. The main reason is the shift between training and test domains~\cite{shift}. In the presence of the domain shift, the model trained on the synthetic data often tends to be biased towards the source domain (synthetic images), making them incapable to generalize to the target domain (real images). 

%In this paper, one of the appeal of Gradient Reversal Loss is the ability to enable the model to learn the domain-invariant representations by punishing the overfitting on the source domain. 

Traditional approaches for domain adaptation mainly focus on the image classification task, which can be summarized as two lines: (1) minimizing the distance between the source and target distributions~\cite{MMD2,MMD1,MMD}; (2) 
explicitly ensuring that two distributions close to each other by adversarial learning~\cite{adaa,CoGAN}. 
Existing works~\cite{DANN,uda} used the similar idea, i.e., gradient reversal layer, to our proposed loss in the domain adaptation for image classification, which was achieved by multiplying a negative scalar during the backpropagation.  
%However, since there exist large category discrepancies between pixels in one image, the manner of uniformly reversing the gradients for all pixels with same scalar is not suitable for the structured prediction in the segmentation. 
%Besides, the loss weight of each pixel should vary to its probability of ground truth class. Different probabilities hold different contributions to the gradient. 
Those drawbacks limit the gradient reversal layer to generalize to the segmentation adaptation. 
%Specially, we thus propose to introduce a novel loss function to dynamically model the importance of each pixel based on the probability of ground truth class. In addition, the Conservative Loss also penalizes the extremely bad or good cases while encouraging the moderate cases.

\begin{figure*}[t]
\begin{center}
   \includegraphics[width=1.0\linewidth]{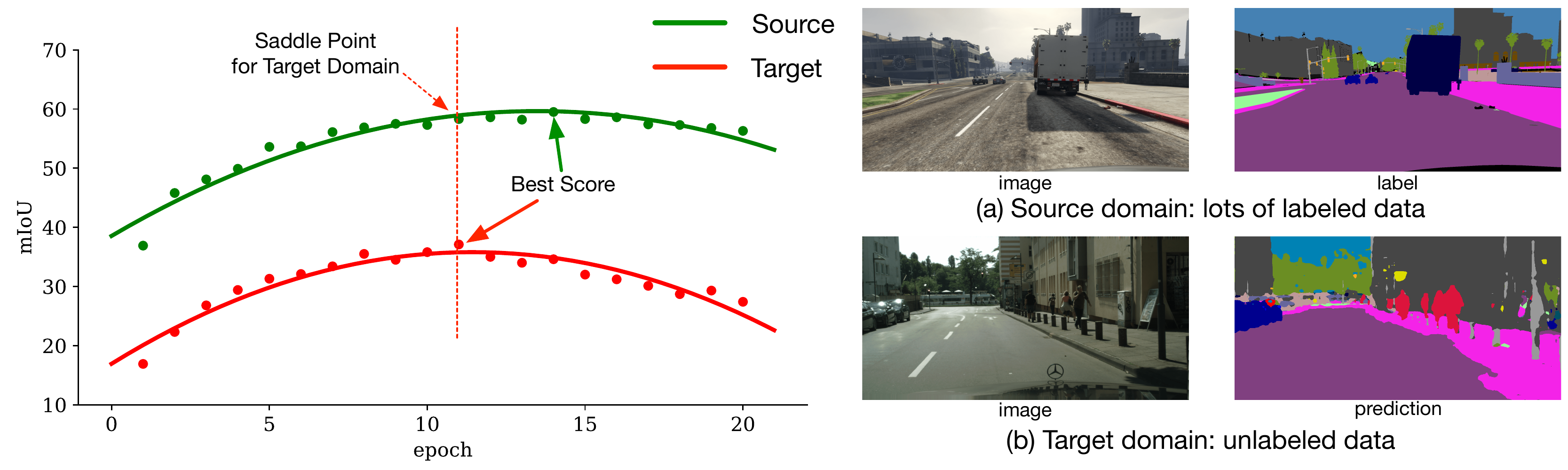}
\end{center}
%\vspace{-4ex}
   \caption{We show the tendency of mIoU on the source domain and target domain. The curves indicate the trends and points denote the actual mIoU. Besides, we display the samples from source domain (GTAV) and target domain (Cityscapes)}
\label{fig:overfit}
%\vspace{-3ex}
\end{figure*}

%In addition, learning to segment an image is more difficult than classification because it has to make the pixel-level predictions in an exponentially large label space and its prediction is highly structured. 
Semantic segmentation provides pixel-level label for input image, which carries more dense and structured information than image classification, and thus making its domain adaptation difficult.
Hence, the domain adaptation techniques in the classification task which focus on sparsely high-level features do not translate well to the segmentation adaptation~\cite{cda}.
Few works have explored the domain adaptation for segmentation~\cite{cda,FCNwild,cross}. Orthogonal to those works focusing on manipulating the data statistics~\cite{FCNwild} or applying the curriculum learning~\cite{cda} to adaptation, we propose the novel Conservative Loss to realize it without introducing extra computational overhead.
%which apply the statistics of data or curriculum learning for adaptation, we propose the novel Conservative Loss to realize it.

%Specially, we thus propose the Gradient Reversal Loss for explicitly assigning different values to different pixels in a natural way.
%since the prediction of segmentation are made in an expon ,  
%However, as pointed out in~\cite{cda}, the domain adaptation techniques in the classification task does not translate well to the semantic segmentation adaptation because its predictions are in an exponentially label space. 
%In this paper, we focus on the segmentation adaptation task with a harder scenario: strong supervision is available in the source domain (synthetic images), but no supervision is available in the target domain (real-world images). 
%extends the range in which an example receives low loss. 
% prove that two properties.

We observe that with training step goes by, the performance on the target domain \emph{first rises and then falls}. We show the trends of mIoU on the experiment of synthetic (GTAV data~\cite{play}) to real (Cityscapes data~\cite{cityscapes}) segmentation adaptation in Fig~\ref{fig:overfit}. It can be observed that the performance on source domain and target domain would not reach the best at the same time because of the domain shift. Since there is no ground truth for target domain during training, it is required to find the saddle point of target domain on the source domain. It is note-worthy that the saddle point for target domain does bias to the best score on the source domain but not reach, which delivers a balance between the discriminativeness and domain-invariant. This phenomenon is consistent with many domain adaptation theories~\cite{the1,the2,the3}.
%when the performance on the target domain achieves the best, the performance on the source domain does not get the best 
%Since the trend on source domain is influenced by the overfitting problem~\cite{overfit}, we argue that this overfitting also degrades the performance of adaptation on target domain.
Therefore, we focus on learning representations with two following characteristics which are: ($\text{{\romannumeral1}}$) discriminative for semantic segmentation on the source domain (corresponding to the `\emph{first rises}') and ($\text{{\romannumeral2}}$) invariant to the change of domains. 
%(escaping from the overfitting).
%Many theoretical analyses of domain adaptation~\cite{the1,the2,the3} have offered a bound on the expected risks of target domain, which depends on its source domain error (test-time) and the divergence measure between two domains. 
%In another word, the features performing well on the target domain in our task hold two following properties which are: ($\text{{\romannumeral1}}$) discriminative for semantic segmentation on the source domain and ($\text{{\romannumeral2}}$) invariant to the change of domains. We thus focus on learning representations which keep those characteristics. 

In this paper, this is achieved by training with the Conservative Loss in an adversarial framework. The Conservative Loss is extremely simple. It holds two attributes corresponding to the properties of desired representations. First, when the probability of ground truth label on the source domain is low, the Conservative Loss enforces the network to learn more discriminative features via gradient descent, which corresponds to the first property of discriminativeness. Second, when the probability of ground truth label is much high, our loss penalizes this case by giving a negative value, which prevents the model from biasing to source domain training data further increasing the generalization capability. This corresponds to the second property of domain-invariant. Our loss function can be seen to seek the optimal parameters that deliver a saddle point of those two objectives. Furthermore, the generative adversarial network (GAN)~\cite{GAN} is also introduced to our model. Unlike some works~\cite{adaa,FCNwild} where they apply the feature-level discriminator, we utilize the GAN to further supplement the domain alignment by enforcing reconstructed images to be indistinguishable for the discriminator. 

We conduct extensive experiments on synthetic to real segmentation adaptation. The proposed method considerably improves over previous state-of-the-art and achieves \textbf{9.3} points of mIoU gain on Synthia~\cite{synthia} to Cityscapes~\cite{cityscapes} experiment without introducing any extra computational overhead during evaluation. Ablation studies verify the effect of different components to our performance and give more insights into properties of our Conservative Loss. 
More discussions and visualization demonstrate the Conservative Loss has good flexibility rather than limiting to a fixed instantiation. 
%More discussions and visualization demonstrate that (1) the Conservative Loss has good scalability rather than restricting an exact form and (2) our model effectively generalizes to semi-supervised setting.

%Below, we first give a brief introduction of related work in Section~\ref{related}. Then we  detail the overall framework and Gradient Reversal Loss in Section~\ref{methods}. Experimental results and analyses are presented in Section~\ref{experiment}.

\section{Related Work}
\label{related}
%1. semantic segmentation and synthetic to real;
%2. DA for image classification
%3. style transfer
%4. gradient reversal or loss reversal, a similar idea but totally different tasks
%5. DA for segmentation

\noindent \textbf{Semantic Segmentation.}~~~
Semantic segmentation is a highly active field, which is a task of assigning object label to each pixel of image. With the surge of deep segmentation model~\cite{FCN}, most recent top-performing methods are built on the CNNs~\cite{PSP,WIDE,decnet}. 

Huge amount of human effort is required to annotate the fined-grained semantic segmentation ground truth. According to~\cite{play}, it did take about 60 minutes to manually segment each image. On the contrary, collecting data from video games such as GTAV~\cite{play} is much faster and cheaper compared with the human annotator. For example,~\cite{play} extracted 24,966 GTAV images with annotations within 49 hours by using a GPU parallel method. However, it is hard to apply the model trained on the synthetic image to the real-world image because of their discrepant data distributions.

%\vspace{2ex}
\noindent \textbf{Domain Adaptation.}~~~
Many machine learning methods rely on the assumption that the training and test data are in the same distribution. However, it is often the case that there exists some discrepancies~\cite{the1,the3}, which leads to significant performance drop on the test data. Domain adaptation aims to alleviate the impact of the discrepancy between training and test data. 

%\vspace{2ex}
\emph{Domain Adaptation for Image Classification.}~~~
Existing works on domain adaptation mostly focus on image classification problem. Conventional methods include Maximum Mean Discrepancy (MMD)~\cite{MMD2,MMD1,MMD}, geodesic flow kernel~\cite{GFK}, sub-space alignment~\cite{SA}, asymmetric metric learning~\cite{sml}, \emph{etc}. Recently, domain adaptation approaches aim to improve the adaptability of deep neural networks~\cite{MMD2,uda,auto,meet,dba,drcn,open,unified}. 
%have been widely explored to improve the adaptability for the domain adaptation. 
%As mentioned above, an idea related to ours is described in~\cite{uda}, in which the gradient reversal layer has similar effect to our proposed loss function, but its task and implementation are quite different (building an image classifier rather than a pixel-level segmenter, and directly reversing the gradient rather than using a loss function).

%\vspace{2ex}
\emph{Domain Adaptation for Semantic Segmentation.}~~~
Much less attention has been given to domain adaptation for semantic segmentation task. The pioneering work in this task is~\cite{FCNwild}, which combines the global and local alignment methods with a domain adversarial training. Another work~\cite{cda} applies the curriculum learning to solve the domain adaptation from easy to hard. In~\cite{cross}, they propose an unsupervised learning to adapt road scene segmenters across different cities. In~\cite{adapt}, they perform output space adaptation at feature level by an adversarial module. 
Unlike them constraining the distribution~\cite{FCNwild} or the output of the network~\cite{adapt}, we propose the Conservative Loss to naturally seek the discriminative and domain-invariant representations. 
%Unlike them, we propose the Conservative Loss to seek the representations

%\vspace{2ex}
\noindent \textbf{Adversarial Learning.}~~~
Recently, Generative Adversarial Network (GAN)~\cite{GAN} has raised great attention. Some works extend this framework for domain adaptation. CoGAN~\cite{CoGAN} achieves the domain adaptation by generating cross-domain instances. Domain adversarial neural networks~\cite{DANN} consider adversarial training for suppressing domain biases. In~\cite{adaa}, they incorporate adversarial discriminative setting to help mitigate performance degradation. In our work, we also incorporate the GAN into our model, whose discriminator drives the source image towards the target one for promoting domain alignment.

%1. semantic segmentation

\section{Methodology}
\label{methods}
\begin{figure*}[ht]
\begin{center}
   \includegraphics[width=1.0\linewidth]{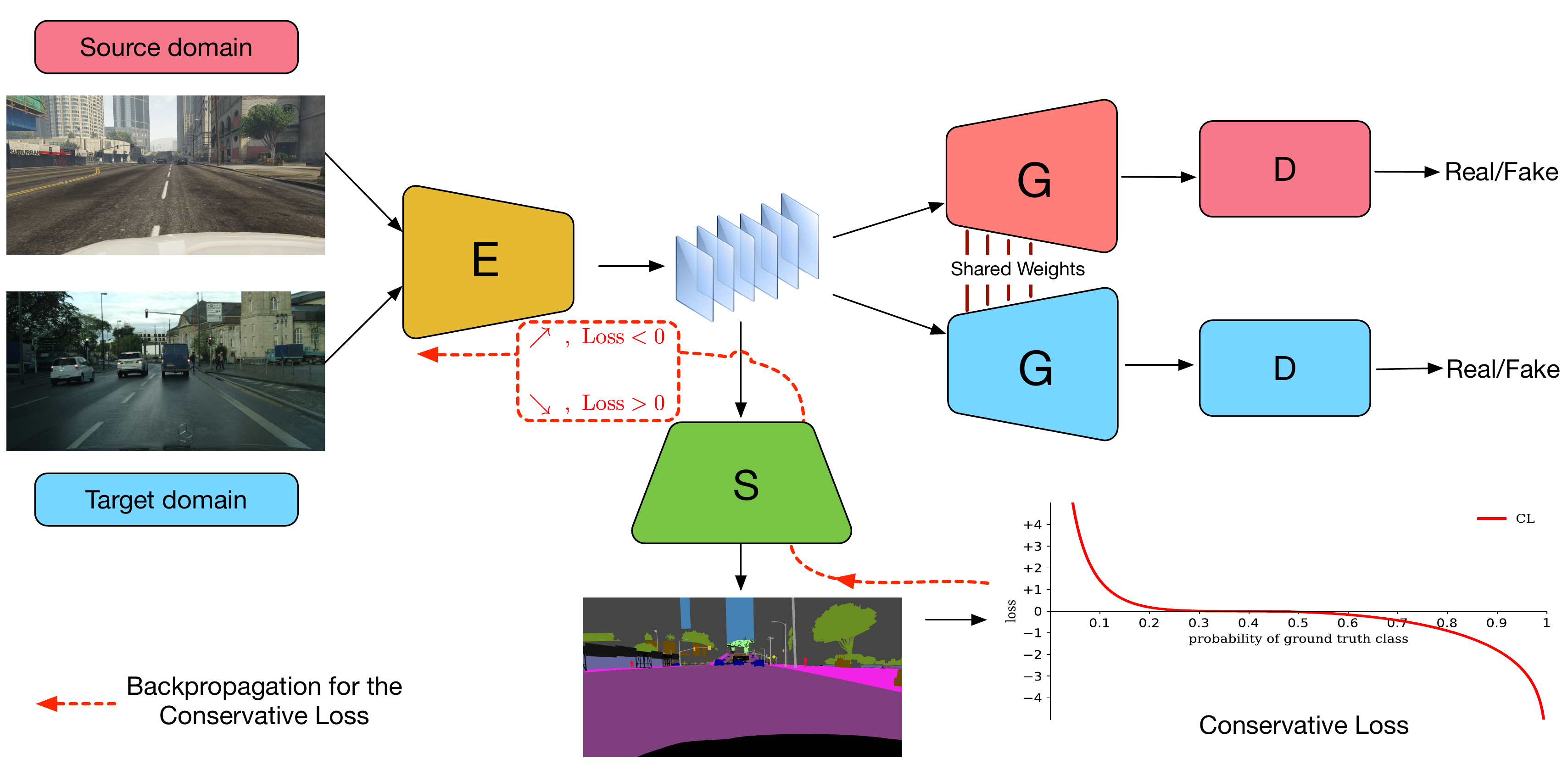}
\end{center}
%\vspace{-4ex}
   \caption{The pipeline of our framework. E denotes the encoder, G denotes the generator, and D is the discriminator. S is the pixel-wise classifier for semantic segmentation. The red color represents the network blocks for the source domain, and the blue for the target domain. We also display the Conservative Loss and its backpropagation. \textcolor{red}{$\nearrow$} represents the gradient ascend and \textcolor{red}{$\searrow$} denotes the gradient descend}
\label{fig:pipeline}
%\vspace{-1ex}
\end{figure*}

As presented above, the key to realize unsupervised domain adaptation is the discriminative and domain-invariant representations. 
The Conservative Loss is proposed to penalize the extreme cases and its goal is to deliver a balance between the discriminative and the domain-invariant representations.
%seek optimal parameters that meet those two characteristics. 
Furthermore, we introduce the generative adversarial networks to align the source and target embedding.
Below, we first describe the framework of our model and its network blocks. Then, the Conservative Loss and its background are presented in details. Finally, the alternative optimization is provided.

\subsection{Framework Overview}
Our framework is illustrated in Figure~\ref{fig:pipeline}.
In our setting, there are two domains: source domain (image and label) and target domain (image only). Our framework aims to achieve good performance on the target domain by applying the model trained on the source domain. 
%During training, we iteratively optimize all three learnable parts (Encoder, Generator and Discriminator). During inference, only the encoder and segmentation classifier are used to produce the results on target domain.
%The GANs enable the model to learn the domain-invariant representations because the encoder enforces the feature embeddings for two domains closer to fool the discriminators.

%\subsection{Network Blocks}

Our model consists of two major parts, i.e., GAN and Segmentation part. The GAN aims to align the source and target embedding. More specifically, the generator and discriminator are playing a minimax game~\cite{GAN}, in which the generator takes source embedding as input and generates the target-like image to fool the discriminator, while the discriminator tries to classify the reconstructed image~\cite{adaa,CoGAN,Zhu2018GenerativeAF}. 
The segmentation part can be seen as a regular segmentation model.  
For each part, the detailed components are shown in the following:
\begin{itemize}
\item[$\bullet$] The encoder(E) performs the feature embedding given source or target image, whose architecture is a fully convolutional network. The generator(G) reconstructs the image based on the embedding. The discriminator(D) does classify the reconstructed images as real or fake. S is the pixel-wise classifier.
\item[$\bullet$] The GAN consists of encoder, generator and discriminator. 
%More specifically, the encoder performs the feature embedding for two domains. The generator reconstructs the image given the embedding from encoder. 
\item[$\bullet$] The segmentation part consists of encoder and pixel-level classifier. Note that the encoder does work in both GAN and Segmentation.
\end{itemize}
The detailed architecture of generators and discriminators is described in the supplementary material because of the limited page space.

%%%%%%%%%%%%%%%%%%%%%%%%%%%%%%%%%%%

\subsection{Background}
\label{back}
In this section, we briefly introduce the theory of domain adaptation and present its relation to our proposed loss.

Many theoretical analyses of domain adaptation~\cite{the1,the2,the3} have offered a upper bound on the expected risks of target domain, which depends on its source domain error (test-time) and the divergence between two domains. Formally, 
\begin{equation}
\begin{split}
\epsilon_{\mathcal{T}} \le \epsilon_{\mathcal{S}} + \frac{1}{2}d(\mathcal{S}, \mathcal{T}) + \mathcal{C},
\end{split}
\end{equation} 
where $\mathcal{S}$ and $\mathcal{T}$ denote the source domain and target domain, respectively. $\epsilon$ is the expected risk. $d$ is the domain divergence, which has different notions, for example $\mathcal{H}$-divergence~\cite{the3}. $\mathcal{C}$ is a constant term.

It can be observed that two terms $\epsilon_{\mathcal{S}}$ and $d(\mathcal{S}, \mathcal{T})$ closely relate to the properties in the desired representations. The first term $\epsilon_{\mathcal{S}}$ indicates that the model should produce discriminative representations for getting smaller expected risks on the source domain, which corresponds to the first property of discriminativeness. The second term $d(\mathcal{S}, \mathcal{T})$ defines the discrepancy distance between two distributions, in which the more similar the representations of both domains are, the smaller it is. This correlates with the second property of domain-invariant. More theoretical analyses are shown in the supplementary material.

%It can be observed that two terms $\epsilon_{\mathcal{S}}$ and $d(\mathcal{S}, \mathcal{T})$ closely relate to the properties in our proposed loss. The first term $\epsilon_{\mathcal{S}}$ indicates that the model should produce discriminative representations for getting smaller expected risks on the source domain, which corresponds to the first property of our loss. The second term $d(\mathcal{S}, \mathcal{T})$ defines the discrepancy distance between two distributions, in which the more similar the representations of both domains are, the smaller it is. This correlates with the second property. More theoretical analyses are shown in the supplementary material.

%%%%%%%%%%%%%%%%%%%%%%%%%%%%%%%%%%%%%%%

\begin{figure*}[ht]
\begin{center}
   \includegraphics[width=0.85\linewidth]{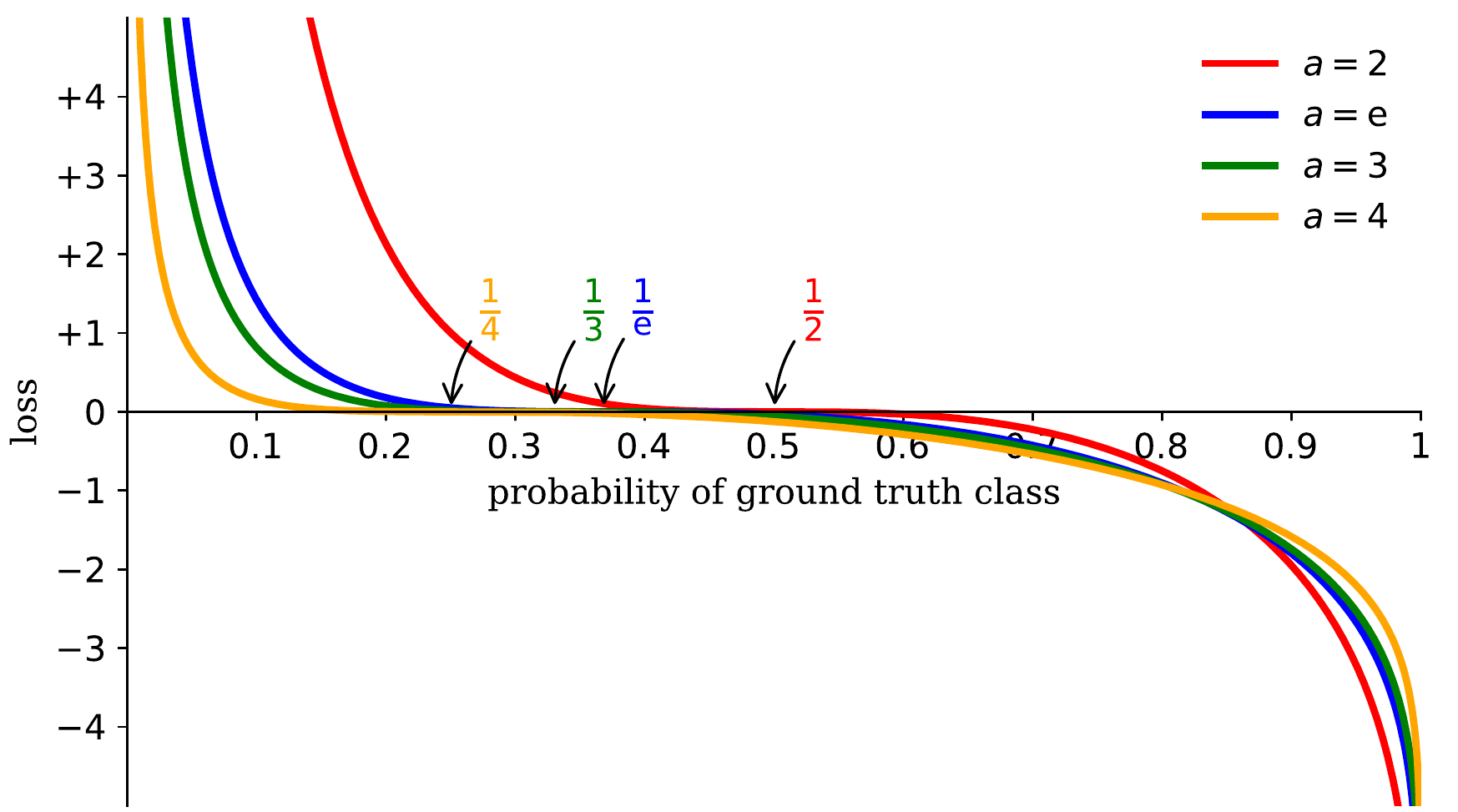}
\end{center}
%\vspace{-4ex}
   \caption{The proposed Conservative Loss with different $a$. It can be observed that the Conservative Loss keeps low values in the middle level and punishes the extremely good or bad cases}
\label{fig:loss}
%\vspace{-3ex}
\end{figure*}

\subsection{Conservative Loss}
\label{reversed}
As explained above, the desired representations should be discriminative for the main task on source domain and possess good generalization ability rather than getting into the overfitting.
%, which significantly degrades the performance on the target domain and source test set (Note that there also exists the distribution mismatch between the source training set and the source test set). 
We thus propose the Conservative Loss for the semantic segmentation on the source domain, which carries the two following properties:
\begin{itemize}
\item[$\bullet$] When the probability of ground truth class is low, the loss function gives a positive value, which enables the network to learn a more discriminative feature by using gradient descent method.
\item[$\bullet$] When the probability is high, the loss function delivers the negative value, which makes the network avoid the bias towards the source domain via the gradient ascend further learning the better generalization. 
\end{itemize}
The Conservative Loss is formulated as:
\begin{equation}
\begin{split}
\text{CL}(p_t) = (1+\log_{a}(p_t))^2 * \log_a({-\log_{a}(p_t)}), \label{eq1}
\end{split}
\end{equation} 
where $p_t$ is the probability of our prediction towards ground truth. $a$ is the base of logarithmic function, which also indicates the intersection point with x-axis, that is $\frac{1}{a}$. The Conservative Loss is visualized for several values of $a\in [2, \mathrm{e}, 3, 4]$ in Figure~\ref{fig:loss}, in which $\mathrm{e}$ is Euler's number and $\mathrm{e}\approx 2.718$.
%for the design of different components
Specifically, $(1+\log_{a}(p_t))^2$ acts as a modulating factor, which delivers the large values when $p_t$ is much low or high. $\log_a(-\log_a(p_t))$ is designed as the switch of gradient direction, in which when $p_t>\frac{1}{a}$ it is negative, otherwise it is positive. 

%Intuitively, the Conservative Loss penalizes the extreme cases 
% need to further refine this line
%Based on the properties of our loss, we introduce several lemmas to further explain the its effect and contribution.

In the following, we have raised two lemmas to analysis the appealing property of our Conservative Loss.

\noindent \textbf{Lemma 1:}~~~\emph{The objective function of domain adaptation system contains a saddle point, which relates to the zero point of Conservative Loss.}
%, which is given by the zero point of Conservative Loss
 
%As the pipeline in Fig~\ref{fig:pipeline} shown, the full objective of the Encoder consists of two parts, including the loss for Segmentation $\mathcal{L}_{seg, p_t}^s$ and the loss for Discriminator $\mathcal{L}_{GAN}$. The brief form of our objective is given by $\mathbb{E} =\mathcal{L}_{GAN} + \mathcal{L}_{seg, p_t}^s$, in which the sign of $\mathcal{L}_{seg, p_t}^s$ dynamically depends on $p_t$. 
As the pipeline in Fig~\ref{fig:pipeline} shown, the full objective consists of two parts, including the loss $\mathcal{L}_{seg, p_t}^s$ for Segmentation  and the loss $\mathcal{L}_{GAN}$ for GAN . The sign of $\mathcal{L}_{seg, p_t}^s$ dynamically depends on $p_t$. 
%When $p_t$ is much high, it is the case the discriminativeness outweighs the domain-invariant. Our loss punishes it via gradient ascend method.
%, and the zero point acts as the saddle point
When $p_t$ is much high, the negative value leads to the gradient ascend for escaping the bias to source domain. Otherwise, the positive value makes the features discriminative. 
It can be seen that our loss balances the two objectives (discriminativeness and domain-invariant) that shape the representations during learning, and its zero point acts as the saddle point. More details are shown in the supplementary material.

\noindent \textbf{Lemma 2:}~~~\emph{Our loss encourages the moderate examples in large range, which makes the overall optimization more stable.}
%Our loss penalizes the extreme cases and encourages the moderate examples.

From the loss form, it can be observed that the loss focuses on the hard negatives and positives, and tends to give the low value for the probability in the middle level. For instance, with $a = \mathrm{e}$, the loss values of $p_t = 0.9$ and $p_t = 0.1$ are -1.8 and 1.4, respectively, while the loss values of $p_t = 0.5$ and $p_t = 0.6$ are -0.03 and -0.06. In such setting, the loss extends the range in which an example receives low loss, which brings a stable optimization even in the case of the gradient descend and ascend frequently alternate due to the joint optimization of $\mathcal{L}_{seg, p_t}^s$ and $\mathcal{L}_{GAN}$.  
%extends the range in which an example receives low loss

In practice we use a $\lambda$-balanced variant of the Conservative Loss:
\begin{equation}
\begin{split}
\text{CL}(p_t) = \lambda(1+\log_{a}(p_t))^2 * \log_a({-\log_{a}(p_t)}).
\end{split}
\end{equation} 
As our experiments will show, different balanced factors $\lambda$ yield slightly different performance. While in our main experiments we use the Conservative Loss defined above, its exact form is not crucial. In Section~\ref{sec:dis} we offer other forms of our loss which also maintain the two properties, and experimental results demonstrate that they can also be effective.

\subsection{Model Objective}

Our full objective is to alternatively update the three network blocks, i.e., discriminators(D), generators(G) and encoder(E). Note that S is a pixel-level classifier which has no learnable parameters in our model. Hence, the objective contains three terms: $\mathcal{L}_{D}$, $\mathcal{L}_G$ and $\mathcal{L}_E$. We then explain the various losses used in our method and describe the alternative optimization scheme.

%\vspace{2ex}
\noindent \textbf{Adversarial Loss.}~~~Inheriting from GAN~\cite{GAN}, we apply the adversarial losses which are derived from the discriminator to all three blocks. We term them as $\mathcal{L}_{\text{GAN,D}}, \mathcal{L}_{\text{GAN,G}}$ and $\mathcal{L}_{\text{GAN,E}}$. For each adversarial loss it consists of two parts, i.e., $\mathcal{L}_{GAN}^s$ for the source image and $\mathcal{L}_{GAN}^t$ for the target image. Thus we can obtain the adversarial loss by $\mathcal{L}_{GAN} = \mathcal{L}_{GAN}^s + \mathcal{L}_{GAN}^t$. 
It is noted that for the encoder, the adversarial loss does a cross-domain update (i.e., classifying the image as real or fake from source domain to target domain and vice versa), which enforces the network to generate similar embeddings for two domains. 
%for the discriminator and generator, the adversarial loss performs the within-domain update (i.e., classifying the image as real or fake in one domain such as source to source or target to target domain).  

%\vspace{2ex}
\noindent \textbf{Reconstructed Loss.}~~~The generator performs the image reconstruction. We use L1 distance as $\mathcal{L}_{rec}$ because L1 encourages less blurring. 

%\vspace{2ex}
\noindent \textbf{Segmentation Loss.}~~~
As Section~\ref{reversed} introduced, the Conservative Loss is applied to the semantic segmentation in the domain adaptation setting. 

%Training of the model results in solving a mini-max problem where the optimization aims to find a saddle point. 
During training, we iteratively optimize all three learnable parts (Encoder, Generator and Discriminator). During inference, only the encoder and segmentation classifier are used to produce the results on target domain. The alternating update scheme is described as following:
\begin{itemize}
\item[(1)] Update discriminators: the overall loss is $\mathcal{L}_D = \mathcal{L}_{GAN,D}$.
\item[(2)] Update generators: the loss involves the adversarial loss and reconstructed loss. The overall loss is $\mathcal{L}_{G} = \mathcal{L}_{GAN, G} + \mathcal{L}_{rec}$.
\item[(3)] Update encoder: since the encoder does work in both two components, i.e., GAN and Segmentation, the overall loss is a combination of several losses, including adversarial loss and segmentation loss on source domain; $\mathcal{L}_{E} = \mathcal{L}_{GAN, E} + \mathcal{L}_{seg}^s$.
\end{itemize}

\section{Experiments}
\label{experiment}

\subsection{Dataset}

Following previous works~\cite{cda,FCNwild}, we use GTAV~\cite{play} or Synthia~\cite{synthia} dataset as the source domain with pixel-level labels, and we use Cityscapes~\cite{cityscapes} dataset as the target domain. We briefly introduce the datasets as following:
%\vspace{-3ex}

\noindent \textbf{GTAV} has 24,966 urban scene images rendered by the gaming engine GTAV. 
%The resolution of images are 1914 $\times$ 1052 and 
The semantic categories are compatible with the Cityscapes dataset. We take the whole GTAV dataset with labels as the source domain data.
%\vspace{-3ex}

\noindent \textbf{Synthia} is a large dataset which contains different video sequences rendered from a virtual city. We take SYNTHIA-RAND-CITYSCAPES~\cite{synthia} as the source domain data which provides 9,400 images from all the sequences with Cityscape-compatible annotations. Inheriting from existing methods~\cite{cda}, we take 16 common object categories for the evaluation.
%\vspace{-3ex}

\noindent \textbf{Cityscapes} is a real-world image dataset focused on the urban scene, which consists of 2,975 images in training set and 500 images for validation. The resolution of images is 2048 $\times$ 1024 and 19 semantic categories are provided with pixel-level labels. We  take the \textbf{unlabeled training set} as the target domain data. The adaptation results are reported on the validation set.

\subsection{Training Setup}
\label{training}
In our experiments, we use the FCN8s~\cite{fcn8s} as the semantic segmentation model. The backbone is VGG16~\cite{vgg16} which is pretrained on the ImageNet dataset~\cite{imagenet}. We apply the PatchGAN \cite{patchgan} as the discriminator, in which the discriminator tries to classify whether overlapping image patches are real or fake. Similar to EBGAN \cite{ebgan}, we add the Gaussian noise to the generator. During training, Adam \cite{adam} optimization is applied with $\beta_1$=0.9 and $\beta_2$=0.999. For the Conservative Loss, we apply $a = \mathrm{e}$ and the balanced weight $\lambda = 5$. The ablation study will give more detailed explanations. Due to the GPU memory limitation, the images used in our experiments are resized and cropped to 1024$\times$512 and the batch size is 1.  More experimental settings will be available in the supplementary material.  
%\vspace{-3ex}

\noindent \textbf{Warm Start.} In our experiments, two different training strategies are employed, which are cold start and warm start. The cold start is that the whole model is trained by using the Conservative Loss from scratch. The warm start indicates the model is trained by first using cross entropy loss and then using our Conservative Loss. Many works~\cite{warm1,warm2,warm3} demonstrate that the warm start strategy to gradient update provides a more stable training compared with cold start.
As the ablation study will show, the warm start performs better than the cold start. In the next domain adaptation experiments, the model is trained using warm start strategy for fairness.

\subsection{Results}
In this section, we provide a quantitative evaluation by performing two adaptation experiments, i.e., from GTAV to Cityscapes and from Synthia to Cityscapes. We compare our method with several existing models, including FCNWild~\cite{FCNwild}, CDA~\cite{cda} and~\cite{adapt}. FCNWild~\cite{FCNwild} applies the dilated network~\cite{dilated} as the backbone and the base model of \cite{cda} is the FCN8s-VGG19~\cite{fcn8s}. Tsai~\textit{et al.}~\cite{adapt} adopts adversarial learning in the output space to perform feature adaptation. The detailed results of each category are available in the supplementary material.

\begin{table}
%\vspace{-3ex}
%\normalsize
%\tiny
%\setlength{\tabcolsep}{0.5pt}
\begin{center}
\caption{Results of domain adaptation from GTAV $\rightarrow$  Cityscapes. The bold values denote the best scores in the column.}
\label{tab:gtav}
%%\vspace{2ex}
\begin{tabular*}{0.72\linewidth}{p{2.5cm}<{\centering}|p{2.5cm}<{\centering}|p{1.5cm}<{\centering}|p{1.8cm}<{\centering}}
\hline
Methods & Base & mIoU & mIoU gain\\
%\rotatebox{90}{road} & \rotatebox{90}{sidewalk} & \rotatebox{90}{building} & \rotatebox{90}{wall} & \rotatebox{90}{fence}& \rotatebox{90}{pole} & \rotatebox{90}{t-light} & \rotatebox{90}{t-sign} & \rotatebox{90}{veg} & \rotatebox{90}{terrain} & \rotatebox{90}{sky} & \rotatebox{90}{person} & \rotatebox{90}{rider}& \rotatebox{90}{car} & \rotatebox{90}{truck}& \rotatebox{90}{bus} & \rotatebox{90}{train}& \rotatebox{90}{mbike} & \rotatebox{90}{bike}& \rotatebox{90}{mIoU}\\
\hline
NoAdapt~\cite{FCNwild} & DilatedNet~\cite{dilated} & 21.1&\\
%\hline
FCNWild~\cite{FCNwild}& DilatedNet~\cite{dilated}  & 27.1 & 6.0\\
\hline
%\hline
NoAdapt~\cite{cda} & FCN8s~\cite{FCN} & 22.3 & \\
%\hline
CDA~\cite{cda} & FCN8s~\cite{FCN} & 28.9 & 6.6\\
\hline
%\hline
Tsai~\textit{et al.}~\cite{adapt} & FCN8s~\cite{FCN} & 35.0 & $-$ \\
\hline
Ours-NoAdapt & FCN8s~\cite{FCN} & 30.0 &\\
Ours & FCN8s~\cite{FCN} & \textbf{38.1} & \textbf{8.1}\\
\hline
\end{tabular*}
\end{center}
%\vspace{-5ex}
%\end{table}
%
%\begin{table}
%The results of 16 common object categories are reported as mIoU and the results of 13 categories (excluding wall, fense and pole) are reported as mIoU-2
%%\vspace{-3ex}
\begin{center}
\caption{Results of domain adaptation from Synthia $\rightarrow$ Cityscapes.}
\label{tab:syn}
%\vspace{2ex}
\begin{tabular*}{0.85\linewidth}{p{2.5cm}<{\centering}|p{2.5cm}<{\centering}|p{1.5cm}<{\centering}|p{1.8cm}<{\centering}|p{1.5cm}<{\centering}}
\hline
Methods & Base & mIoU & mIoU gain& mIoU-2 \\
\hline
NoAdapt~\cite{FCNwild} & DilatedNet~\cite{dilated} & 17.4 & &  \\
%\hline
FCNWild~\cite{FCNwild}&DilatedNet~\cite{dilated}& 20.2 & 2.8 &  \\
\hline
%\hline
NoAdapt~\cite{cda} & FCN8s~\cite{FCN} & 22.0 & & \\
%\hline
CDA~\cite{cda} & FCN8s~\cite{FCN} & 29.0 & 7.0 &\\
\hline
Tsai~\textit{et al.}~\cite{adapt} & FCN8s~\cite{FCN} & $-$ & $-$ & 37.6\\
\hline
Ours-NoAdapt & FCN8s~\cite{FCN} & 24.9 & &\\
Ours & FCN8s~\cite{FCN} & \textbf{34.2} & \textbf{9.3} & \textbf{40.3}\\
\hline
\end{tabular*}
\end{center}
%\vspace{-1ex}
\end{table}

\noindent \textbf{GTAV $\rightarrow$ Cityscapes.} For a fairness, the result is evaluated over the 19 common classes. From Table~\ref{tab:gtav} shown, our proposed method achieves the best performance (mIoU=\textbf{38.1}), which has \textbf{9.2} points higher than~\cite{cda} and \textbf{11} points higher than~\cite{FCNwild}. 
Due to the different experimental settings and backbone network (baseline method \cite{cda} also mentions the difference), our own baseline performance is higher than other methods. However, the highlight is the \textbf{performance gain}. We can find that the proposed method yields an improvement of \textbf{8.1} points higher than 6.0 in~\cite{FCNwild} and 6.6 in~\cite{cda}.
%It is noted that the proposed model gets almost best scores

%%\vspace{-2ex}
%\vspace{-3ex}
\noindent \textbf{Synthia $\rightarrow$ Cityscapes.} We report the results of mIoU in Table~\ref{tab:syn}. It is noted that~\cite{adapt} reported the results on Synthia~\cite{synthia} to Cityscapes adaptation with only 13 object categories (excluding wall, fense and pole). We also report this results as the mIoU-2.
Our proposed model achieves a mIoU of \textbf{34.2}, and more importantly our model obtains a \textbf{9.3} points of performance gain which is higher than the performance gain of \cite{cda} (7.0) and~\cite{FCNwild} (2.8). Compared with~\cite{adapt} on 13 categories, our method also achieves the better performance. In particular, our model does not use any additional scene parsing data except the  source domain and target domain data, while the~\cite{cda} uses another dataset, i.e., PASCAL CONTEXT dataset, to obtain the superpixel label.

\begin{table}
%\vspace{-9ex}
%\small
\setlength{\tabcolsep}{7.0pt}
\centering
\caption{Results of ablation study for different components in the proposed model. CL means the Conservative Loss. CE means the cross entropy loss}
\label{tab:ablation}
%\vspace{2ex}
\begin{tabular*}{0.82\linewidth}{l|c|c|c}
\toprule
Model & FCN8s+CE & FCN8s+GAN+CE & FCN8s+GAN+CL\\
\hline
mIoU & 30.0 &34.4 & 38.1\\
\bottomrule
\end{tabular*}
%\footnotetext{haha}
%%\vspace{2ex}
%\vspace{-4ex}
%\end{table}
%
%\begin{table}
%\vspace{-4ex}
%\small
\setlength{\tabcolsep}{7.0pt}
\centering
\caption{Results of ablation experiments for $a$ and $\lambda$ in the Conservative Loss}
%\vspace{2ex}
\begin{tabular*}{0.65\linewidth}{l|c|c|c|c}
\toprule
$a$ (with fixed $\lambda = 5$) & 2 & $\mathrm{e}$ & 3 & 4\\
\hline
mIoU & 37.5 & 38.1 & 37.3 & 36.8\\
\midrule
\midrule
$\lambda$ (with fixed $a = \mathrm{e}$) & 1 & 5 & 10 & 20\\
\hline
mIoU & 37.2 & 38.1 & 37.9 & 37.8\\
\bottomrule
\end{tabular*}
%\footnotetext{haha}
%%\vspace{2ex}
\label{tab:ab2}
%\vspace{-4ex}
\end{table}

\subsection{Ablation Study}
In this section, we perform the thorough ablation experiments, including experiments with different components, different factors in the Conservative Loss and different training strategies. Those experiments demonstrate different contributions of components and provide more insights of our method. 

\noindent \textbf{Effect of different components.}
In this experiment, we show how each component in our model affects the final performance. We consider several cases as following: (1): the baseline model, which contains only the base segmentation model (FCN8s in our model) and is trained using source data only. (2) the FCN8s and GAN component, which consists of base model and GAN and is trained using both source data and target data with the cross entropy loss. (3) the full model, which involves three parts, i.e., base model, GAN and Conservative Loss. We perform the ablation experiments on GTAV$\rightarrow$Cityscapes setting.
% vgg, vgg+gan, vgg+gan+Conservative Loss
% cross-entropy, focal-loss, Conservative Loss
% for the \lambda
% warm start and cold start

The results of ablation study are shown in Table~\ref{tab:ablation}. It can be observed that  each component plays an important role in performance improvement. 
More specifically, our full model achieves the best results and obtains \textbf{8.1} points performance gain. The GAN part also gets 4.4 performance gain compared with FCN8s+CE. 
%We replace the backbone in~\cite{uda} with FCN8s-vgg16~\cite{FCN}. It can be observed that gradient reversal layer in~\cite{uda} also achieves better results than FCN8s+CE, which indicates the gradient reversal mechanism works in the domain adaptation setting.
Note that the GAN component could introduce the unlabeled target domain data into the whole model, so the Conservative Loss is applied based on the GAN and there is no variant of FCN8s+CL.

\noindent \textbf{Effect of $a$ and $\lambda$ in the Conservative Loss.}
In this part, we design the ablation experiments for $a$ and $\lambda$ in the Conservative Loss. As shown in Equation~\ref{eq1}, $a$ is the base of logarithm and denotes the intersection point with x-axis. $\lambda$ is a balanced factor. We show the impacts of different $a$ and $\lambda$ in Table~\ref{tab:ab2}.
%p{3cm}<{\centering}

Since there are two variables, we perform the ablation study for one variable with another fixed. For the ablation of $a$ (with fixed $\lambda = 5$), it can be observed that $a = \mathrm{e}$ achieves the best result. Furthermore, we can find that all different $a$ obtain much better performance compared with the cross entropy loss (34.4 in Table~\ref{tab:ablation}), which demonstrates that our loss performs consistently better and has high robustness. For the ablation of $\lambda$ (with fixed $a = \mathrm{e}$), different $\lambda$ show slightly different results and $\lambda = 5$ obtains the best performance.
%Furthermore, with $a$ getting bigger, the result is getting worse because the 

\noindent \textbf{Warm start $\&$ Cold start.} As described in Section~\ref{training}, we use a warm start strategy to train the proposed model. In this experiment, we compare the two training strategies. For the cold start strategy, we clamp the Conservative Loss with [$\min=-10$, $\max=10$], while this constraint is not exist in the warm start. We use the $\lambda$-balanced Conservative Loss with $\lambda=5$ and $a=\mathrm{e}$.   
\begin{table}
%\vspace{-4ex}
%\small
\setlength{\tabcolsep}{7.0pt}
\centering
\caption{Results of two training strategies, i.e., cold start and warm start. CL means the Conservative Loss}
\label{tab:warm}
%\vspace{2ex}
\begin{tabular*}{0.85\linewidth}{c|c|c|c}
\toprule
Loss Function & \cite{cda} & CL with cold start &  CL with warm start \\
\hline
mIoU & 28.9 & 35.2 & 38.1\\
\bottomrule
\end{tabular*}
%\footnotetext{haha}
%%\vspace{2ex}
%\vspace{-4ex}
\end{table}

In Table~\ref{tab:warm}, it can be observed that the Conservative Loss with cold start outperforms \cite{cda} with a large margin (6.3 points). The warm start performs better than the cold start because it enables the network to train stably.

\subsection{Discussion}
\label{sec:dis}
In this section, we design several experiments to verify the capability of the proposed method. We show the effect of adaptation on distribution to measure how domain gap is reduced in the feature level. We compared with several classification losses and homogeneous losses to show its superiority and flexibility. 
%Furthermore, we generalize our method to semi-supervised learning.

\noindent \textbf{Visualizations.}
\begin{figure*}[ht]
%\vspace{-8ex}
\begin{center}
   \includegraphics[width=0.85\linewidth]{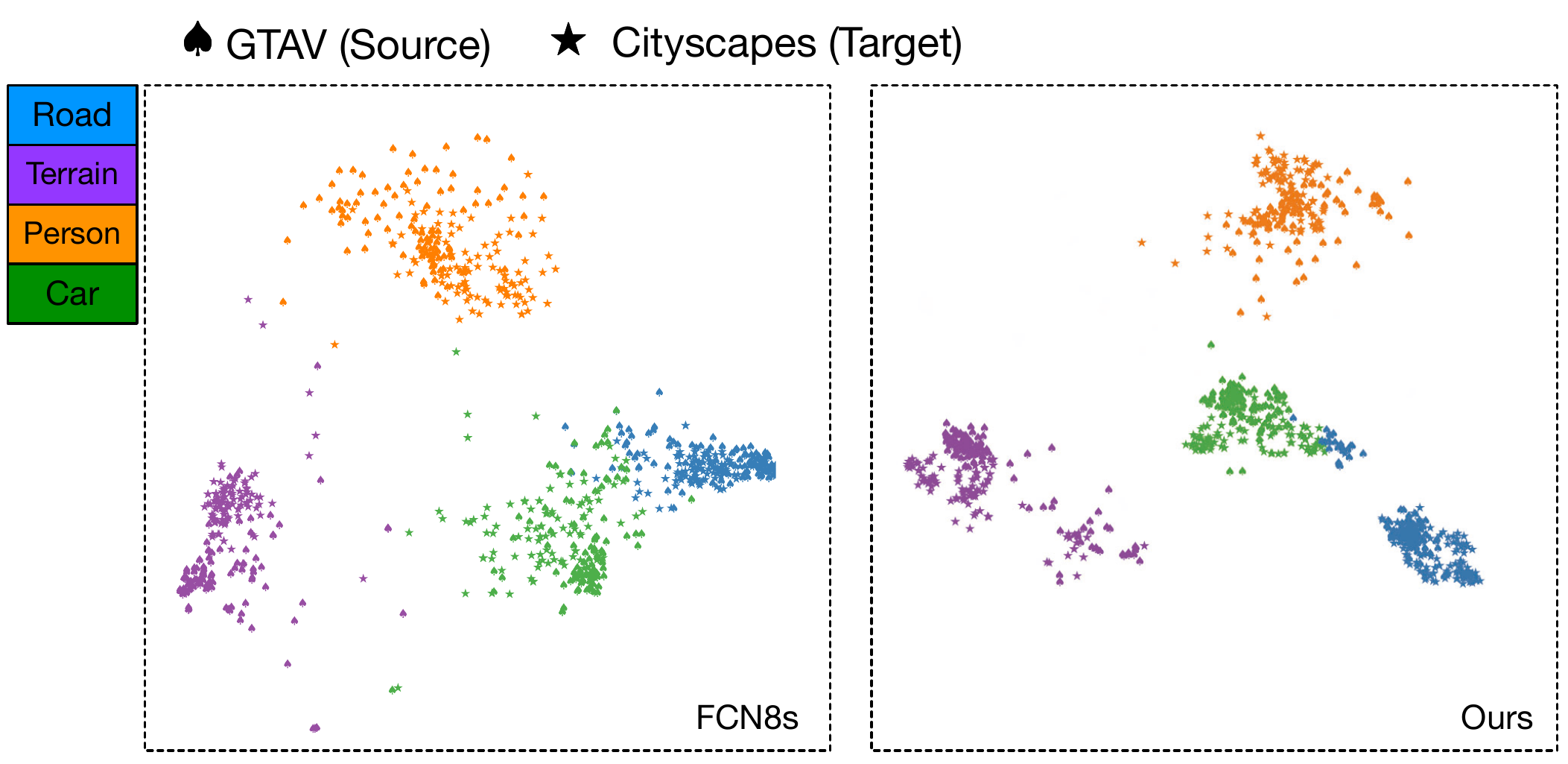}
\end{center}
%\vspace{-3ex}
   \caption{We show the effect of adaptation on the distribution of the extracted features. $\spadesuit$ denotes the point from source domain and $\bigstar$ is from target domain}
\label{fig:tsne}
%\vspace{-4ex}
\end{figure*}
To verify the effect of adaptation on the distribution, we use t-SNE~\cite{tsne} to visualize feature distributions in Figure~\ref{fig:tsne}. 100 images are randomly selected from each domain and for each image the features from last convolutional layer (the channel size equals to class categories.) are extracted. We compare the distributions of our model with FCN8s (No adaptation). Four categories are sampled to display for a clearly visual effect. We observe that with the adaptation applying, the distance between two domains with same class becomes closer and the discrepancy between different classes also gets clear.  

% for x^3, x^seg, etc
% for the source domain with unlabeled target data
%\subsection{}
\begin{figure}[!ht]
%\vspace{-1ex}
    \centering
       \includegraphics[width=0.57\linewidth]{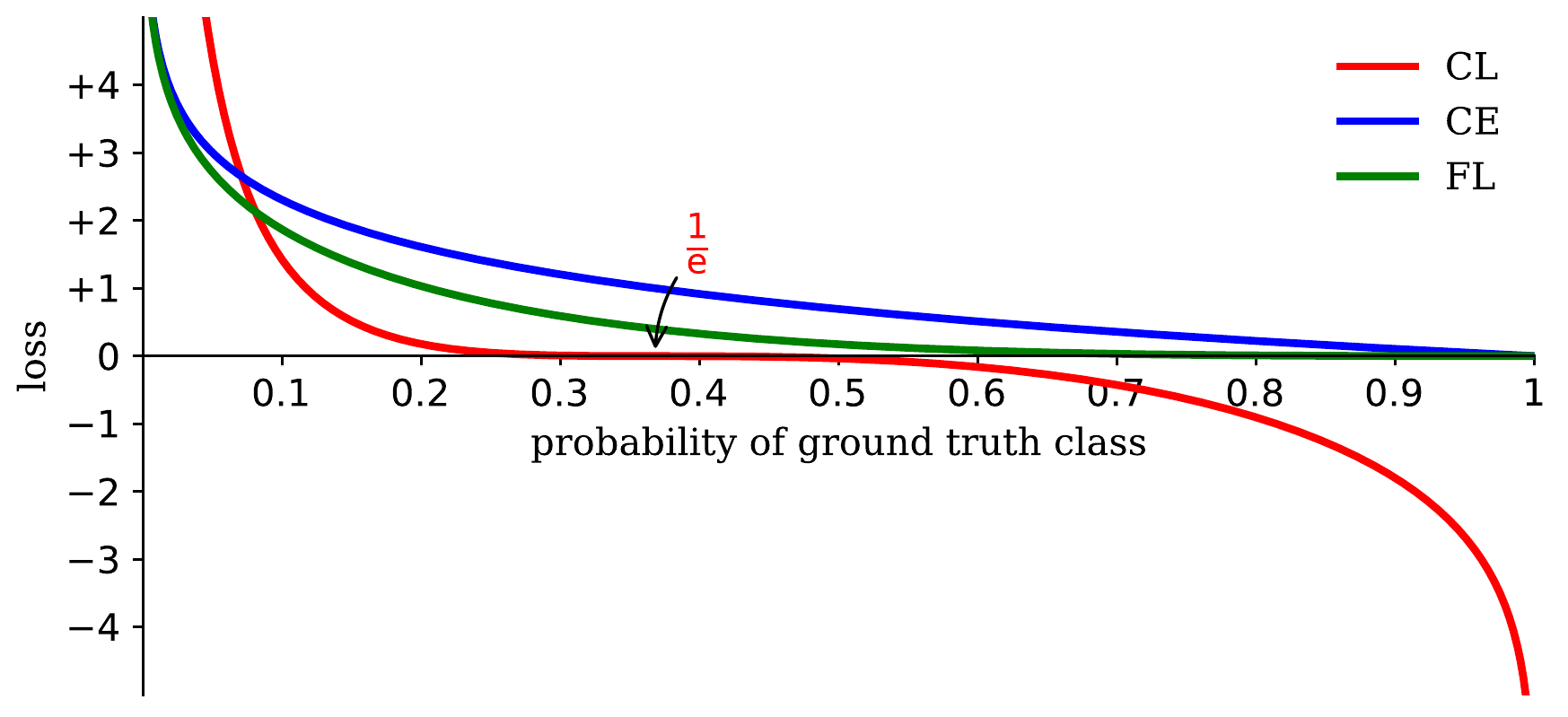}
    \qquad
    \small
    \begin{tabular}[b]{cc}
%    \small
    \toprule
      Loss & mIoU \\
    \hline
    \hline
      Cross Entropy & 34.4\\
    \hline
      Focal Loss & 35.8 \\
    \hline
      Conservative Loss & 38.1\\
    \bottomrule
    %\vspace{1ex}
    \end{tabular}
    %\vspace{-2ex}
    \caption{The left figure shows three classification losses, including Cross Entropy loss (CE) in blue, Focal Loss (FL) in green and Conservative Loss (CL) in red. The right table shows the results of all three losses on GTAV $\rightarrow$ Cityscapes adaptation experiment}
    \label{fig:loss2}
    %\vspace{-3ex}
\end{figure}
\noindent \textbf{Comparison with other classification losses.}
In this experiment, we compare the Conservative Loss to Cross Entropy Loss and Focal Loss~\cite{focal}. The Cross Entropy Loss is given by $\text{CE}(p_t) = -\log(p_t)$, which is plotted in Figure~\ref{fig:loss2} with green line. To ensure fairness, we utilize the $\alpha$-balanced Focal Loss $\text{FL}(p_t) = -\alpha_t(1-p_t)^2\log(p_t)$ and warm start in the experiment of Focal Loss, and apply $\alpha_t = 5$ by using a cross-validation. 

From the right table in Figure~\ref{fig:loss2}, it can be observed that the Focal Loss obtains a better performance compared with the cross entropy loss because it focuses learning on hard negative examples. However, in the domain adaptation, the domain-invariant representations are crucial to achieve good adaptation performance. The Conservative Loss does enable the network to be insensitive to domain changes by punishing the extreme cases. It can be seen that the Conservative Loss yields higher result (\textbf{38.1}), and obtains more performance gain (\textbf{3.7}) than the Focal Loss ({1.4}) based on the cross entropy loss.

\begin{figure*}[h]
%\vspace{-3ex}
\begin{center}
   \includegraphics[width=0.8\linewidth]{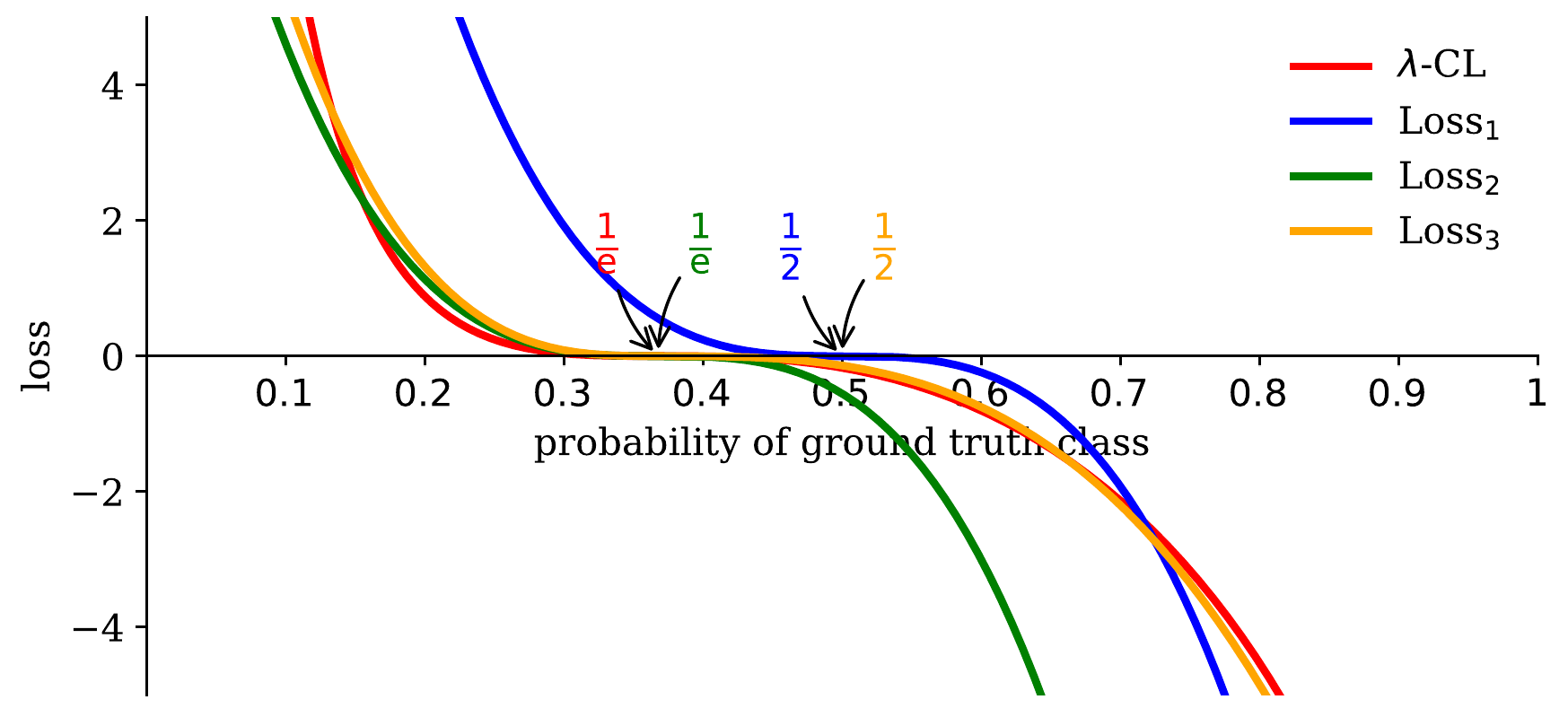}
\end{center}
%\vspace{-3ex}
   \caption{The plot of $\lambda$-balanced Conservative Loss (CL) and several homogeneous losses. Note that the $\lambda$-balanced Conservative Loss is used in the figure, in which the value of Conservative Loss multiplies the balanced factor $\lambda = 5$}
\label{fig:loss3}
%\vspace{-8ex}
\end{figure*}
\noindent \textbf{Effect of homogeneous losses.} As shown in Section~\ref{reversed}, the Conservative Loss has two properties: (1) when the $p_t$ is low, the Conservative Loss enforces the network to learn discriminative features. (2) when the $p_t$ is high, the loss enables the network to learn domain invariant features by gradient ascend method, which aims to penalize the extremely good cases. There are several losses that also maintain these two properties, for example the cubic equation. In this experiment, we propose several homogeneous losses to verify the effect of these two properties, which are given by:
\begin{align}
\text{Loss}_1 &= -\lambda_1(p_t-0.5)^3, \label{eq6} \\
\text{Loss}_2 &= -\lambda_2(p_t-\frac{1}{\mathrm{e}})^3, \label{eq7}\\
\text{Loss}_3 &=\left\{
\begin{array}{rcl}
-\alpha*(p_t - \frac{1}{\mathrm{e}})^3,& &{p_t < \frac{1}{\mathrm{e}}}, \\\\
-\beta*(p_t - \frac{1}{\mathrm{e}})^3,& &{p_t \ge \frac{1}{\mathrm{e}}}. 
\end{array} \right.
\label{eq8}
\end{align}
%\begin{spacing}{1.8}
%\begin{equation}
%%\normalsize
%\text{Loss}_3=\left\{
%\begin{array}{rcl}
%-\alpha*(p_t - \frac{1}{\mathrm{e}})^3,& &{p_t < \frac{1}{\mathrm{e}}} \\\\
%-\beta*(p_t - \frac{1}{\mathrm{e}})^3,& &{p_t \ge \frac{1}{\mathrm{e}}} 
%\end{array} \right.
%\label{eq8}
%\end{equation}

Equation~\ref{eq6} and \ref{eq7} demonstrate the $\lambda$-balanced cubic equations with different intersection points, i.e., 0.5 and $\frac{1}{\mathrm{e}}$, respectively. Equation~\ref{eq8} is a piecewise function, which is more similar to the Conservative Loss due to these two balanced factors. We show these homogeneous losses in Fig~\ref{fig:loss3}.
%The distributions of all these losses are visualized in Figure~\ref{fig:loss3}, in which all these losses possess two properties explained above.

\begin{table}
%\vspace{-2ex}
%\small
\setlength{\tabcolsep}{7.0pt}
\centering
\caption{Results of homogeneous losses}
\label{tab:ablloss}
%\vspace{2ex}
\begin{tabular*}{0.8\linewidth}{c|c|c|c|c|c|c}
\toprule
Loss Function & CE & FL& Loss$_1$ & Loss$_2$ & Loss$_3$ & CL\\
\hline
mIoU & 34.4 & 35.8 & 36.5 &36.7 & 37.8 & 38.1\\
\bottomrule
\end{tabular*}
%\footnotetext{haha}
%%\vspace{2ex}
%\vspace{-4ex}
\end{table}

We apply the adaptation experiment on GTAV $\rightarrow$ Cityscapes to verify their capabilities. The results are reported in Table~\ref{tab:ablloss}. In order to ensure fairness, all experiments are performed based on the warm start and those hyper-parameters ($\lambda_1, \lambda_2, \alpha, \beta$) are chosen by using the cross-validation. We can observe that all homogeneous losses perform better than the cross entropy loss (34.4) and Focal Loss (35.8).
%Furthermore, the piecewise function Loss$_3$ achieves better performance compared to the Loss$_1$ and Loss$_3$. 
Therefore, we can find that the exact form of the Conservative Loss is not crucial, and several homogeneous losses also yield comparable results and perform better than cross entropy loss and Focal Loss. Generally, we expect any loss function with similar properties as Conservative Loss to be equally effective.

\section{Conclusion}
\label{conclusions}
In this paper, we have proposed a novel loss, the Conservative Loss, for the semantic segmentation adaptation. To enforce the network to learn the discriminative and domain-invariant representations, our loss combines the gradient descend and gradient ascend method together, in which it penalizes the extreme cases and encourages moderate cases. We further introduce the adversarial networks to our full model for supplementing the domain alignment. Extensive experiments demonstrate our model achieves state-of-the-art. Exploratory experiments show that the Conservative Loss has high flexibility without limiting to exact form.

\noindent \textbf{Acknowledgments}~~~ This work is partially supported by the Big Data Collaboration Research grant from SenseTime Group (CUHK Agreement No. TS1610626).

\bibliographystyle{splncs04}
\bibliography{egbib}
\end{document}